\documentclass{article}
\usepackage{ijcai20}

\usepackage{times}

\usepackage{soul}
\usepackage{url}
\usepackage[hidelinks]{hyperref}
\usepackage[utf8]{inputenc}
\usepackage[small]{caption}
\usepackage{graphicx}
\usepackage{amsmath}
\usepackage{booktabs}
\usepackage{amsfonts}
\usepackage{multirow}
\usepackage{subfigure}

\title{Transfer Heterogeneous Knowledge Among Peer-to-Peer Teammates: A Model Distillation Approach}

\author{
Zeyue Xue$^1$\footnote{Equal contribution. \quad Correspondence to: Chao Wu, chao.wu@zju.edu.cn, Pan Zhou, zhoupannewton@gmail.com$>$.}\and
Shuang Luo$^{2*}$\and
Chao Wu$^{2}$\and
Pan Zhou$^{1}$\and
Kaigui Bian$^{3}$\And
Wei Du$^4$\\
\affiliations
$^1$Huazhong University of Science and Technology\\
$^2$Zhejiang University\\
$^3$Peking University\\
$^4$University of Arkansas, Fayetteville\\
}

\begin{document}

\maketitle

\begin{abstract}
Peer-to-peer knowledge transfer in distributed environments has emerged as a promising method since it could accelerate learning and improve team-wide performance without relying on pre-trained teachers in deep reinforcement learning. However, for traditional peer-to-peer methods such as action advising, they have encountered difficulties in how to efficiently expressed knowledge and advice. As a result, we propose a brand new solution to reuse experiences and transfer value functions among multiple students via \emph{model distillation}. But it is still challenging to transfer Q-function directly since it is unstable and not bounded. To address this issue confronted with existing works, we adopt Categorical Deep Q-Network. We also describe how to design an efficient communication protocol to exploit heterogeneous knowledge among multiple distributed agents. Our proposed framework, namely Learning and Teaching Categorical Reinforcement (LTCR), shows promising performance on stabilizing and accelerating learning progress with improved team-wide reward in four typical experimental environments.
\end{abstract}

\section{Introduction}

Deep Reinforcement Learning (DRL) has been employed to train autonomous agents to solve complex sequential decision-making problems such as board and video games \cite{mnih2015human}, robotics \cite{kober2013reinforcement}, DRAM memory control \cite{barto2017some} and so on. However, in many complex domains, the learning speed is too slow to be feasible. And the DRL community has devoted much effort to alleviate this burden.  Recently, peer-to-peer knowledge transfer among distributed agents has drawn considerable concern since it can accelerate individual learning speed and improve team-wide performance \cite{omidshafiei2019learning} \cite{kim2019heterogeneous}. Another benefit of learning peer-to-peer is that it can accelerate learning even without relying on the pre-trained teachers. The key remaining question is to develop robust methods to autonomously reuse knowledge and pass on that knowledge to uniformed teammates, which is a process named as \emph{teaching}. There are some previous works on teaching, such as \emph{action advising}: learning when and what actions to advise to a teammate. However, 
the estimate is difficult to obtain since each student would follow different learning trajectories and obtain converged policies finally.
Another challenge is to represent and use the action-value function that most DRL methods learn. Furthermore, a good \emph{action-advising} teacher should not only advise the best action but also guide others to explore when necessary. To address these issues, we propose a brand new solution to explicitly transfer value functions and reuse experiences among peer-to-peer teammates via \emph{model distillation}.

Model distillation \cite{hinton2015distilling} was first presented as an efficient means for supervised model compression, and it has been extended to the problem of creating a single network from an ensemble of models \cite{fukuda2017efficient}. There have been some previous works on applying model distillation to the field of DRL \cite{rusu2015policy} \cite{berseth2018progressive}. But they just considered the scenario where a student learns the ensembled policies from a group of pre-trained experts. Once an expert is not given, it is less clear how to learn the optimal Q-function and policy. However, it is still challenging to directly transfer a Q-function via model distillation since the Q-values are not bounded and can be very unstable. Consequently, we adopt Categorical Deep Q-networks (DQN) \cite{bellemare2017distributional}, which is a brand new form that adds a \emph{softmax} layer to the output probability distribution of action-value function on fixed discrete support, which provides a richer set of predictions and a more stable target for learning. With Categorical DQN on hand, model distillation also has emerged as a good candidate for exploiting distributed heterogeneous knowledge, which is expressed by the predictions of each student's value function. 

It is noted that our method is different from traditional cooperative Multi-Agent Reinforcement Learning (MARL)
such as independent Deep Q-Network \cite{tampuu2017multiagent}, where DQN has been extended to MARL settings, in which each agent observes the joint state, selects an individual action and receives a team reward but could not guarantee its convergence. It is because the environment dynamics are non-stationary from any individual agent’s perspective. However, our method, namely Learning and Teaching Categorical Reinforcement (LTCR), does not use joint action or state space but works well with guaranteed convergence in multiplayer-mode simulated environments.
In this work, a group of teammates learn from each agent’s heterogeneous knowledge and obtain local categorical Q-network. Our method further has the following desirable properties:

\begin{enumerate}
    \item Our method leverages the knowledge contained in a limited number of demonstrations. Knowledge is given by a group of teammates periodically, and each agent utilizes the ensembled knowledge to enhance the learning speed and stability with improved team-wide performance.
    \item Our method is flexible, as it can be improved with most new deep Q-network extensions such as Dueling Networks \cite{wang2015dueling}, Double Q-learning \cite{van2016deep}, Noisy Nets \cite{fortunato2017noisy} and Rainbow \cite{hessel2018rainbow}.
    \item Our scalable method supports heterogeneous agents which own heterogeneous skills and value functions but are with the same reward distribution in single-player and multi-player simulated environments.
\end{enumerate}

\section{Background}
\textbf{Agents and Environments}. 
In deep reinforcement learning, an agent always interacts with the environment discretely to accomplish a task.
At each discrete time step $t=$
$0,1,2 \ldots,$ the environment provides the agent with an observation $s_{t},$ the agent responds by selecting an action $a_{t}$ and then the environment provides the next reward $r_{t+1}$, discount $\gamma,$ and state $s_{t+1}.$ This above interaction is formalized as a Markov Decision Process (MDP), which is a tuple $\langle\mathcal{S}, \mathcal{A}, T, R, \gamma\rangle,$ where $\mathcal{S}$ is a finite set of states, $\mathcal{A}$ is a finite set of actions, 
$r(s, a)=\mathbb{E}\left[r_{t+1}\right|\left.s_{t}=s, a_{t}=a\right]$ is the reward function, and $\gamma \in[0,1]$ is a discount factor. 

Action selection is given by an agent's policy $\pi$
that defines a probability distribution over actions for each
state. And for a fixed policy $\pi$, the return, $R^{\pi}=\sum_{t=0}^{\infty} \gamma^{t} r_{t},$ is a random variable representing the sum of discounted rewards observed along one trajectory of states when following policy $\pi$. Many RL algorithms estimate the action-value function.
\begin{equation}
Q^{\pi}(s, a):=\mathbb{E}\left[R^{\pi}(s, a)\right]=\mathbb{E}\left[\sum_{t=0}^{\infty} \gamma^{t} R\left(s_{t}, a_{t}\right)\right],
\end{equation} and the optimal action-value function, $Q^{*}(s, a) =\max _{\pi}$ $Q^{\pi}(s, a),$ obeys the Bellman optimality equation
\begin{equation} 
Q^{*}(s_{t}, a_{t})=\mathbb{E}_{s_{t+1}}[r_{t+1}+\gamma \max _{a_{t+1}} Q^{*}\left(s_{t+1}, a_{t+1}\right)],
\end{equation} where $s_{t+1}$ is the next state after the state-action pair $(s_{t}, a_{t})$ according to the (stochastic) transition function $P\left[s_{t+1}=s^{\prime} | s_{t}=s, a_{t}=a\right]$.\\ \\
\textbf{Distributional DRL and Categorical DQN.} We can learn to approximate the distribution of value functions $Q^{\pi}(x, a)$. Recently \cite{bellemare2017distributional} has proposed to model such distributions with probability masses placed on a discrete support $\boldsymbol{z},$ where $\boldsymbol{z}$ is a vector with K atoms, defined by $z_{k}=v_{\min }+(k-1) \frac{v_{\max }-v_{\min }}{K-1}, \quad k \in\left\{1, \ldots, K\right\}. $ The atom probabilities are given by a parametric model $\theta: \mathcal{S}\times\mathcal{A}\to R^{K}$.

The approximating distribution $Z^{\theta}(s_{t},a_{t})$ output by a \emph{softmax} is defined on this support, with the probability mass $p^{\theta}_{k}\left(s_{t}, a_{t}\right)$ on each $k,$ such that $Z^{\theta}(s_{t},a_{t})=\left(\boldsymbol{z}, p^{\theta}\left(s_{t}, a_{t}\right)\right)$.
The goal is to update $\theta$ such that this distribution closely matches the actual distribution of returns.

To learn the probability masses, the key insight is that return distributions satisfy a variant of Bellman's equation. 
Categorical DQN is derived by first constructing a new support for the target distribution $Z^{\theta^{\prime}}(s_{t},a_{t})$, and then minimizing the Kullbeck-Leibler divergence between the distribution $Z^{\theta}(s_{t},a_{t})$ and the target distribution, which is given by  $Z^{\theta^{\prime}}(s_{t},a_{t}) \equiv\left(r_{t+1}+\gamma \boldsymbol{z}, \quad \boldsymbol{p}^{\theta^{\prime}}\left(s_{t+1}, a_{t+1}^{*}\right)\right)$, and the loss function is shown as:
\begin{equation}
L(s_{t},a_{t})=D_{\mathrm{KL}}\left(\Pi_{C} Z^{\theta^{\prime}}(s_{t},a_{t}) \|Z^{\theta}(s_{t},a_{t}) \right).
\end{equation}

Similar to traditional DQN, we still use a frozen copy of $\theta^{\prime}$ to construct a target network. $\Pi_{C}$ is a Cramer-projection of the target distribution onto the fixed support $\boldsymbol{z,}$ and $a_{t+1}^{*}=\operatorname{argmax}_{a} z^{\top} p^{\theta^{\prime}}\left(s_{t+1}, a\right)$ is
the greedy action with respect to the mean action  values in state $s_{t+1}$.\\\\
\textbf{Linear Function Approximation}
However, it is still challenging to discuss Categorical DQN with model distillation. For further theoretical analysis, we introduce the linear function approximation. In this setting, we represent each state-action pair $(s_{t}, a_{t})$ as a feature vector $\phi_{s_{t}, a_{t}} \in \mathbb{R}^{d}.$ We wish to find a linear function given by a weight vector $\theta$ such that
\begin{equation}
 Q(s_{t}, a_{t}) \approx \theta^{T} \phi_{s_{t}, a_{t}}.
\end{equation}

And in the categorical distributional setting, $\theta$ becomes a matrix $W \in \mathbb{R}^{K \times d}.$ Here we will consider approximating distribution function on $z_{k}$:
\begin{equation}
Z_{z_{k}}\left(s_{t},a_{t}\right) \approx  W \phi_{s_{t}, a_{t}}[k].
\end{equation}

In this setting, there may be no $W$ for which $Z_{z_{k}}\left(s_{t},a_{t}\right)$ describes a proper distribution function: e.g., Cumulative Distribution Function $F(y)$ of $Z_{z_{k}}\left(s_{t},a_{t}\right)$ may be less than or greater than 1 for $y>z_{K}.$ However, as shown by \cite{bellemare2019distributional}, we can still analyze the behaviour of a distributional algorithm which is allowed to output improper distributions. 
Consequently, we write $\mathcal{Z}:=\left\{W \phi: W \in \mathbb{R}^{K \times d}\right\}$ is the set of distributions of value functions that are linearly representable over $\phi$. For convenience, we will utilize $Z^{\phi}\left(\boldsymbol{z}_{k}\right)$ to represent $Z_{z_{k}}\left(s,a\right)$ in the following.
\section{Overview of LTCR}
\subsection{Distilling to Teach}
Distillation is a method to transfer knowledge from a teacher model $Z_{i}$ to a student model $Z_{j}$. The traditional distillation targets from a classification network are typically obtained by passing the weighted sums of the last network layer through a \emph{softmax} function. Traditional distillation process starts with a powerful teacher network and performs one-way knowledge transfer to an untrained student. However, a pre-trained expert is always unavailable in social settings.

Recently, \cite{zhang2018deep} extended model distillation to a setting where a pool of untrained students which learn simultaneously to solve a classification task. Trained in this way, it turns out teammates learn significantly better than when learning independently. Similarly, in our scenario, the value functions of each untrained student are given by a \emph{softmax} layer. Given N feature vectors $\Phi=\{(s_{n}, a_{n})\}_{n=1}^{N}$ with K atoms,
we adopt the distillation setup of \cite{zhang2018deep} and minimize the Kullback-Leibler (KL) divergence, which is shown as:
\begin{equation}
D_{K L}\left(Z_{i} \| Z_{j}\right)=\sum_{n=1}^{N} \sum_{k=1}^{K} Z^{\phi_{n}}_{i}\left(\boldsymbol{z}_{k}\right) \log \frac{Z^{\phi_{n}}_{i}\left(\boldsymbol{z}_{k}\right)}{Z^{\phi_{n}}_{j}\left(\boldsymbol{z}_{k}\right)}.
\end{equation}

\begin{figure}[ht]
    \centering
    \includegraphics[width=8.5cm]{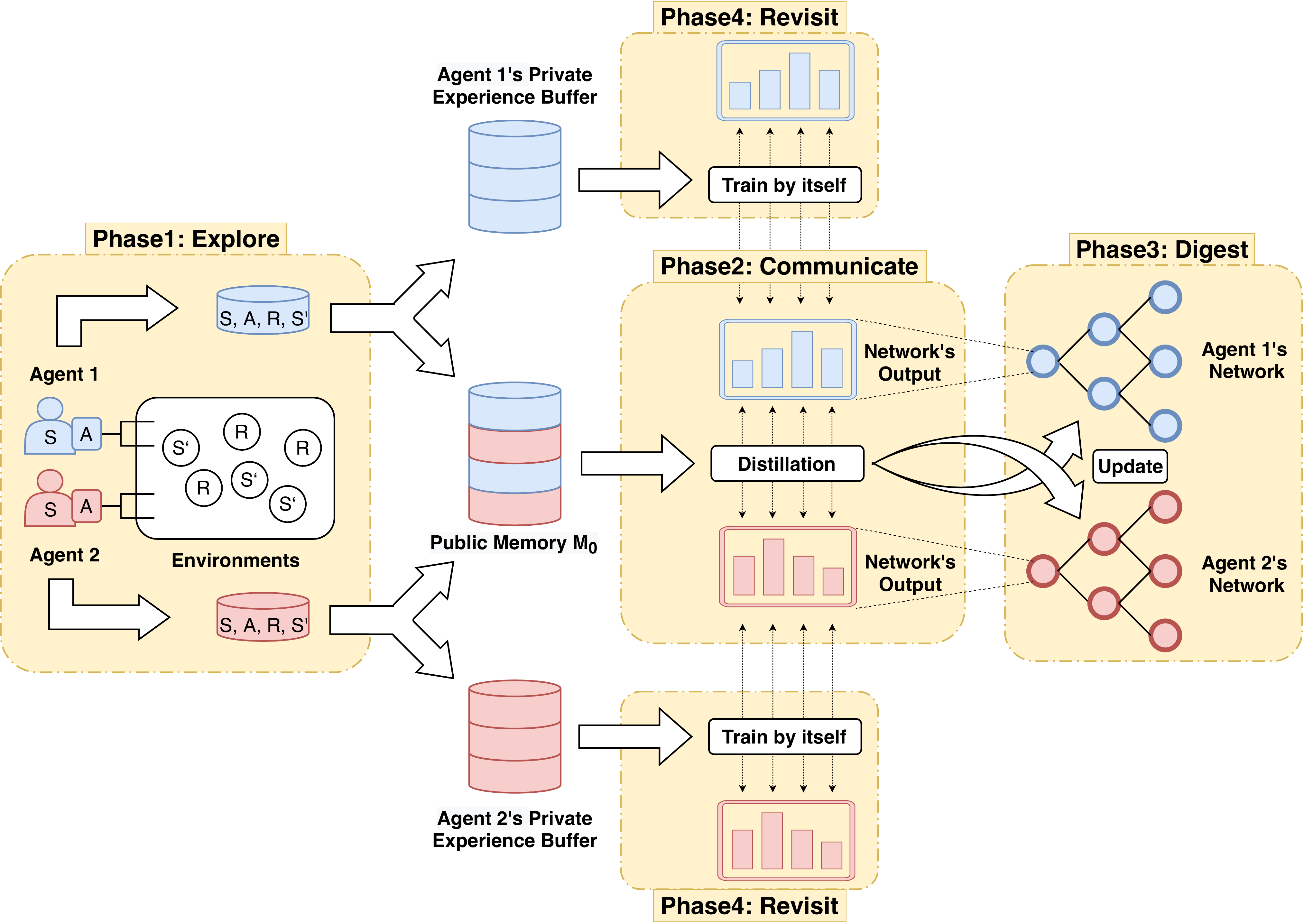}
    \caption{Agents teach each other according to LTCR. Teaching occurs between the teacher value function of one agent and the student value function of another agent.}
    \label{fig:process}
\end{figure}

In our framework, there are two roles: the role of the teacher agent $i$, (i.e., an agent whose value function $Z_{i}$ gives teaching demonstrations) and the role of the student agent $j$, (i.e., an agent who receives teaching demonstrations and enhances its value function $Z_{j}$). Agent $i$ and $j$ are able to teach each other via model distillation, but we only consider a one-way interaction for clarity: agent $j$ is willing to minimize $D_{KL}(Z_{i}\|Z_{j})$. So how does KL-divergence metric work? We have argued that $\mathcal{Z}_{j}:=\left\{W_{j}(t) \phi_{n}: W(t) \in \mathbb{R}^{K \times d}\right\}$ with feature vector $\phi_{n}=(s_{n},a_{n})$. Assume the weight matrix initializes at zero, $W_{j}(0)=\textbf{0}$, and with Theorem 1 in \cite{phuong2019towards}, we could argue that weight matrix $W_{j}(t)$ fulfills almost surely:
$$
W_{j}(t) \to W_{i} \quad t\to \infty,
$$
when we adopt the notion of \emph{gradient flow} (stepsize is finite and sufficiently small). As a result, LTCR solves the problem of transferring the knowledge of categorical Q-function $Z_{i}$. Each agent also owns a private experience replay to minimize $L_{i}(s_{t},a_{t})$, shares a memory device $M_{0}$ to establish a communication protocol and performs model distillation. Our approach of learning involves 4 phases (see Figure \ref{fig:process}):

\textbf{Phase 1: Explore.} Each student explores the environment for storing transitions in private experience buffer and upload feature vectors $\Phi$ to memory $M_{0}$.

\textbf{Phase 2: Communicate.} Each student randomly selects a much smaller subset $\Phi_{0}\in M_{0}$, computes the value function distribution on each vector in $\Phi_{0}$ and communicates with teammates.

\textbf{Phase 3: Digest.} Each student calculates KL-divergence based on teaching demonstrations and perform distillation for $t_{i}$ epochs.

\textbf{Phase 4: Revisit.} Each student trains its model on its private experience buffer for $t_{i+1}$ epochs.

\begin{figure}[ht]
\centering
\includegraphics[width=8cm]{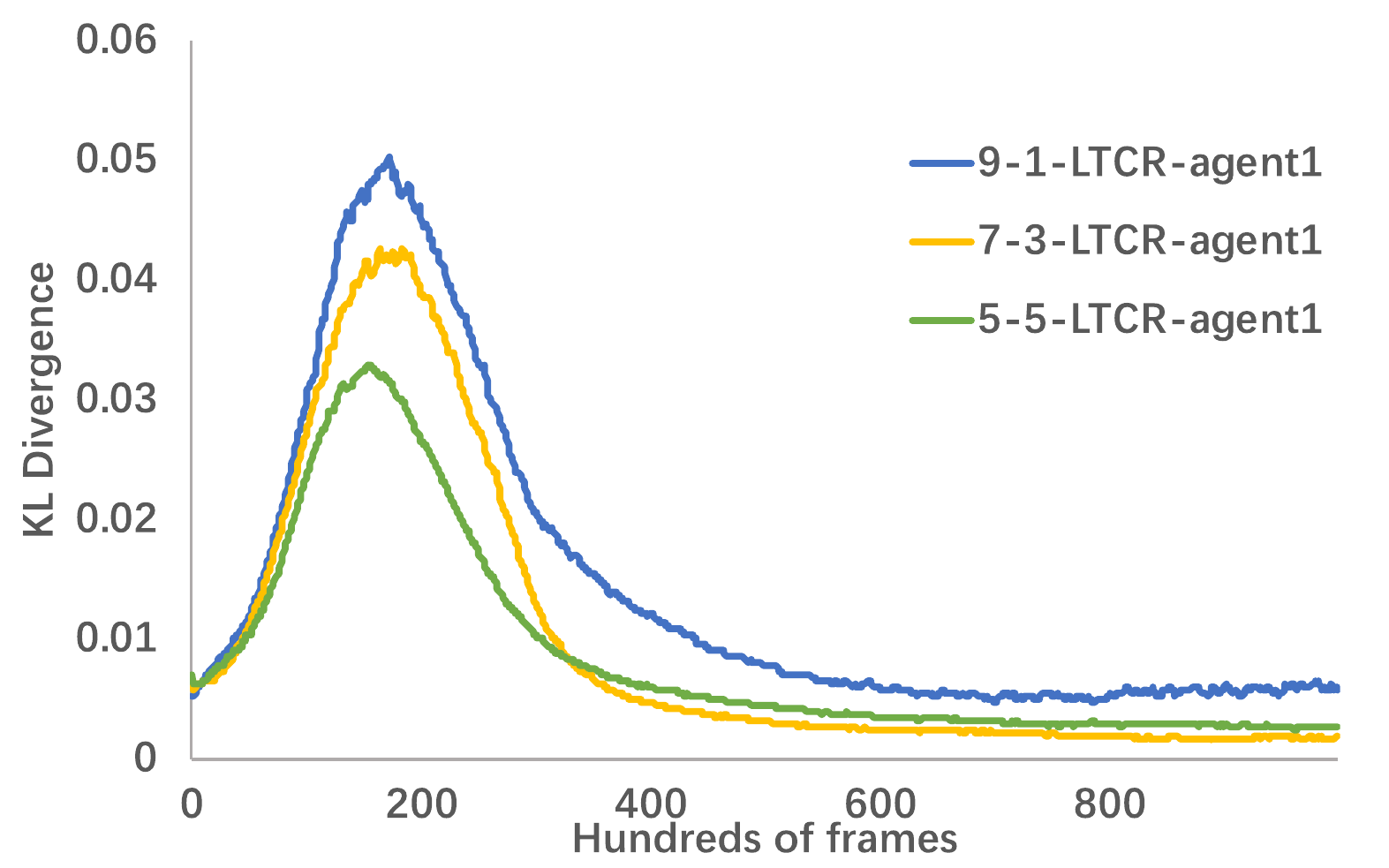}
\caption{We vary the ratio between Revisit steps and Digest steps at 9:1, 7:3, 5:5, namely 9-1-LTCR, 7-3-LTCR, 5-5-LTCR. This figure indicates the tendency of KL-divergence between teammates with the purpose on emphasizing the influence of the KL metric during the training process in a simulated environment Cartpole.}
\label{fig:kl}
\end{figure}

No restrictions are placed on agents algorithms but the same reward distribution (i.e., they can be heterogeneous). Iteration of Phase 2-4 enables
training with two losses: a conventional learning loss $L(s_{t},a_{t})$ , which generally enhances local performance, and a mimicry loss $D_{KL}$, which transfers knowledge among teammates. These two losses result in increasingly capable of the optimal categorical Q-function and policy. Figure 2 depicts the tendency of KL-divergence. It indicates that $D_{KL}$ increases in the early stage mainly due to the conventional loss $L(s_{t},a_{t})$, and decreases to almost zero in the later stage because of model distillation.

\subsection{Communication}
\subsubsection{Knowledge Heterogeneity}
During the process of knowledge transfer, knowledge heterogeneity among distributed agents is likely to arise.
The underlying stochastic transition process naturally causes each agent to explore different sections of the environment. Agent $i$ also sometimes enters a foreign state its teammates have already mastered. Consequently, each private experience buffer would contain unique experiences, develop different skills, and accumulate heterogeneous knowledge, which results in heterogeneous categorical deep Q-network $Z_{i}$.
\subsubsection{Communication Protocol}
Consequently, model distillation is considered as a promising solution to transfer heterogeneous knowledge among distributed teammates. So how to establish efficient communication protocol to perform model distillation? The main challenge is to leverage the knowledge contained in a limited number of demonstrations. 
\begin{figure}[ht]
\centering
\subfigure[Cartpole]{
    \includegraphics[height=3cm]{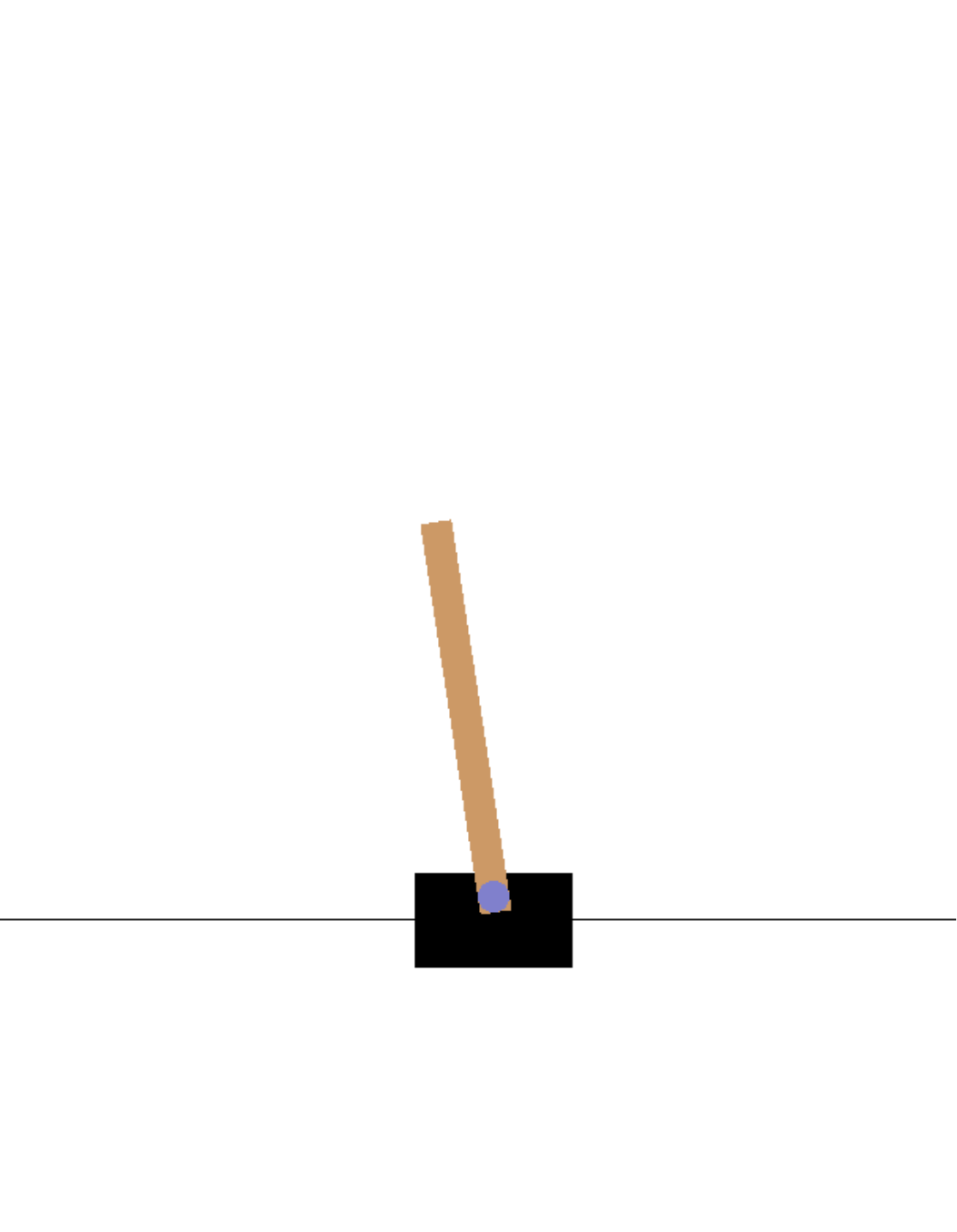}
}
\subfigure[Space Battle]{
    \includegraphics[height=2.7cm]{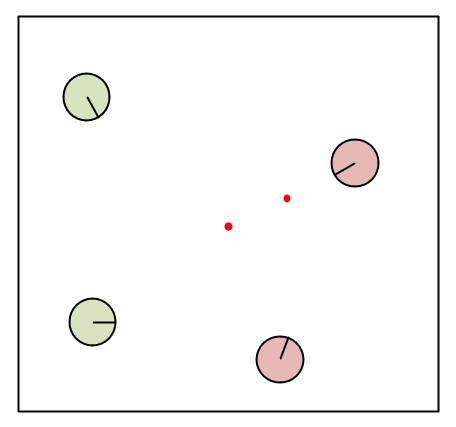}
}
\subfigure[Zaxxon]{
    \includegraphics[height=3cm]{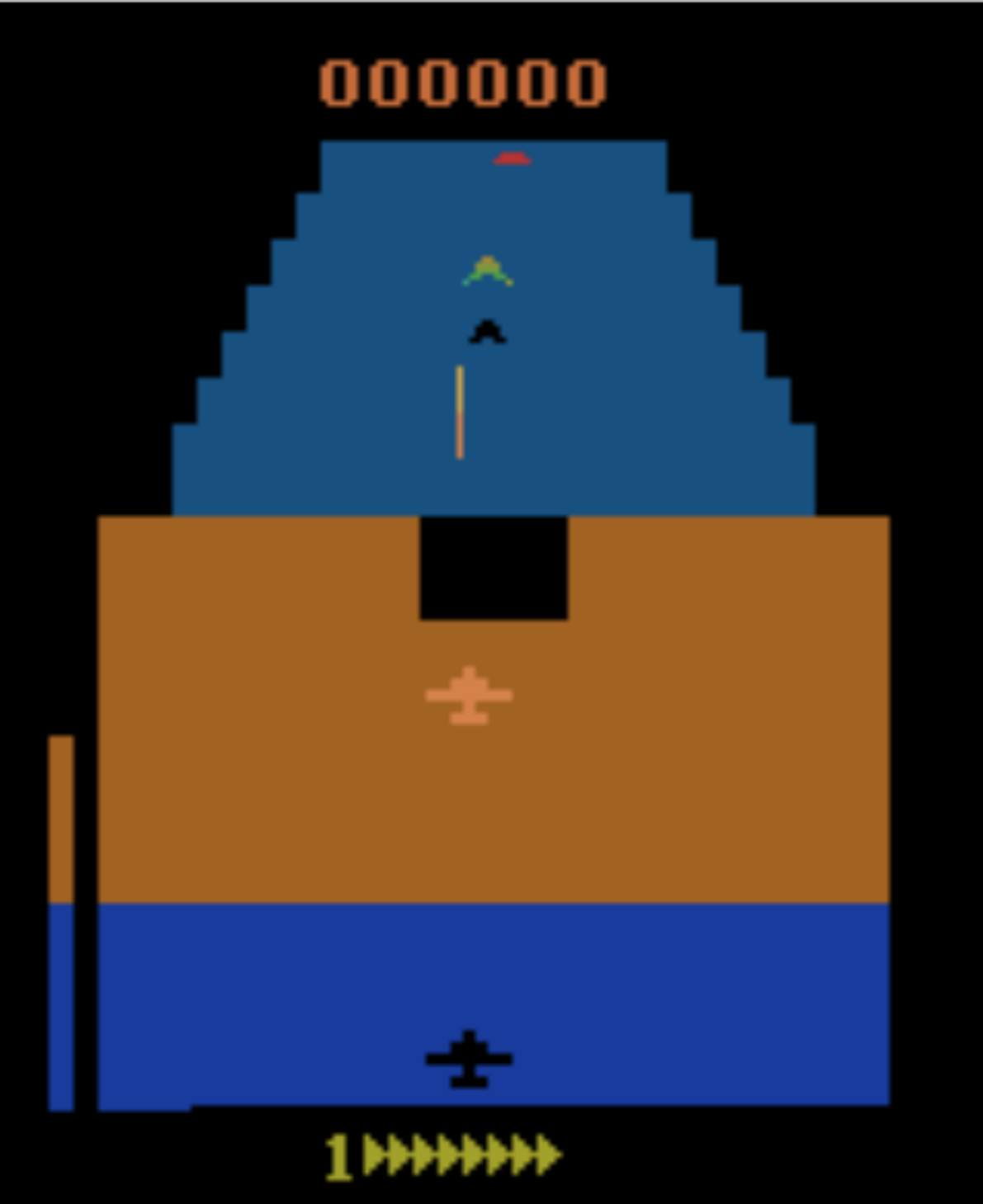}
}
\caption{Visualization of three environments. }
\label{fig:spacebattle}
\end{figure}
In the one-way interaction process,
we utilize the public memory $M_{0}$ as the basis of communication, which is realized through knowledge distillation. 
Teacher expresses its knowledge by sharing the value function distributions at sampled feature vectors $\phi_{n}$. In this way, the knowledge of the teacher can be understood by others without explicitly sharing its model architecture and private experience buffer. In each communication round, demonstrations are typically 
 recorded as a vector including feature vectors $\Phi=\{(s_{n}, a_{n})\}_{n=1}^{N}$ (where multiple state features are composed to describe a state $s_{n}$) and value function distributions on these feature vectors. Then student agent verifies these vectors in $M_{0}$ and calculate KL-divergence at these feature vectors to perform model distillation.

In the traditional model distillation settings, each student always owns a non-i.i.d. labeled dataset. Consequently, teammates have the same target labels to perform gradient descent. However, in the context of DRL, each student independently designs its target network to minimize $L(s_{t},a_{t})$, which provides an unstable target for collaboration. As a result, for example, the soft-target could be designed as the average of each original target distributions to enhance the stability of training process.

\section{Experiments}
In this section, we discuss our experimental environments, explain our methodology, and then show our main result. 
For our study, we use Categorical deep Q-network as \cite{bellemare2017distributional} our baseline. We perform our environment in both single-player and multi-player modes for analysis, and we provide all environments details below.

\subsection{Experiment Setup}

\subsubsection{Single-player mode}
To measure the performance and robustness of our framework in a variety of environments, we test LTCR in both simple environments such as Cartpole and more complex environments belonging to Arcade Learning Environment.

Cartpole is a game where a pole is upright to a cart, and the goal is to prevent it from falling over (see Figure \ref{fig:spacebattle}(a)). The system is controlled by applying a force of pushing a cart to the left or right, and a reward of +1 is provided for every timestamp that the pole remains upright. The maximum reward is 200, while if the pole is more than 15 degrees from vertical or the cart moves more than 2.4 units from the center, the game ends. 
\begin{figure}[ht]
\centering
\includegraphics[width=8cm]{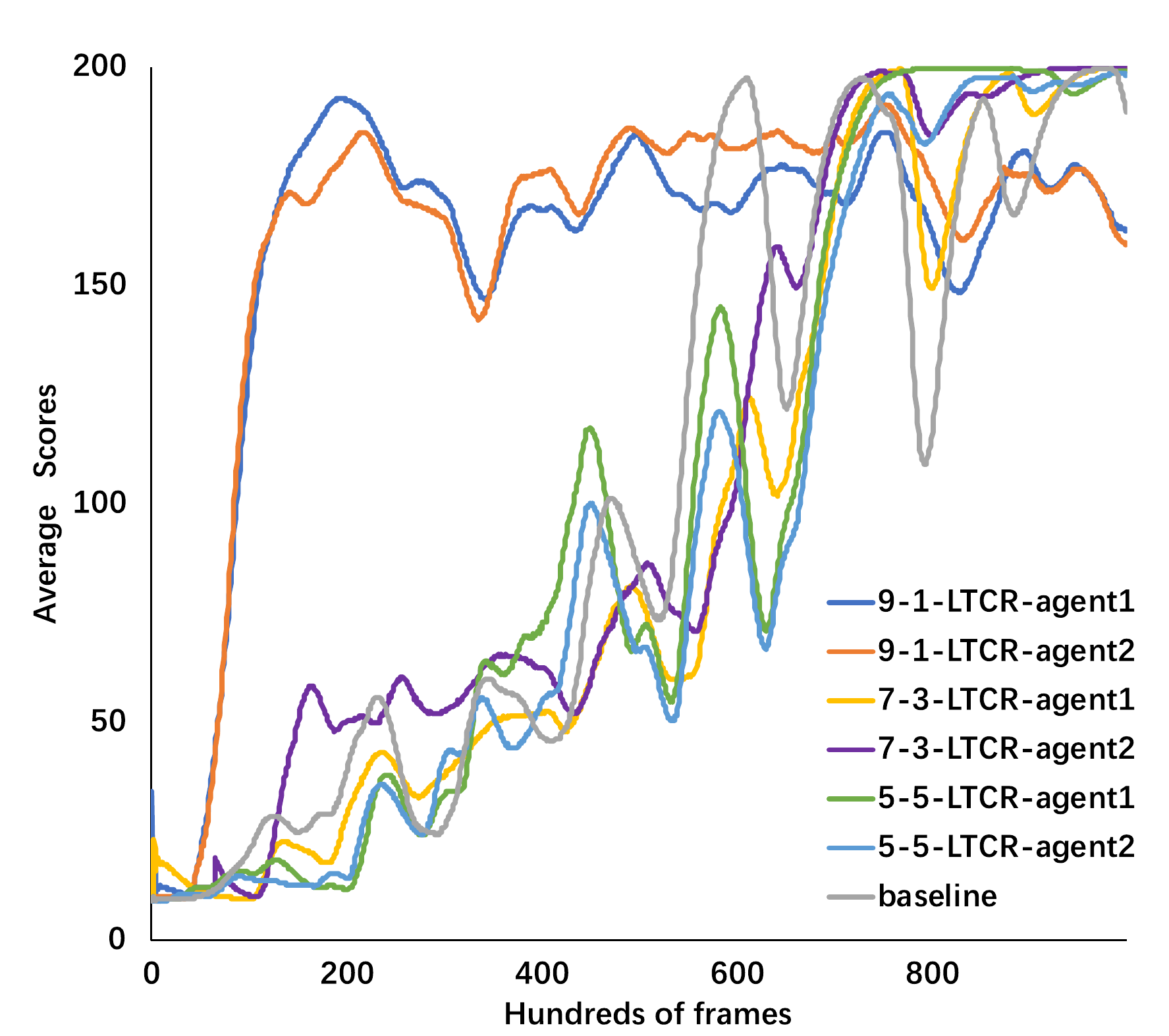}
\caption{This figure compares the learning curves of different kinds of LCTR, representing different ratios between Revisit steps and Digest steps in Cartpole.}
\label{fig:cartpoleReward}
\end{figure}
The state space is represented as a 4-tuple vector, indicating the position and velocity information of the cart as well as the angle and velocity at the tip of pole.

Zaxxon and Phoenix are both the games on Atari 2600 Games. Zaxxon is the game that scrolls diagonally through enemy fortresses and outer space as the player's ship dodges the shots of enemy fighters and force field barriers, finally encountering the evil armored robot Zaxxon(see Figure \ref{fig:spacebattle}(c)). Phoenix is an outer space-themed, fixed shooter video game released in arcades in 1980. Each game's state space is represented by a 84*84 tuple vector, and the action space size is 18. The reward distribution originates from the game's original settings.

\subsubsection{Multi-player mode}
Space battle is a simulated ship team game (see Figure \ref{fig:spacebattle}(b)). To perform our experiments, we adopt the grounded communication environment, inhabiting a two-dimensional world with approximately continuous space, discrete time and discrete action space. There are two teams in our design. Agents of each team could attack enemy by launching ammunition or hitting each other directly to reduce the remaining blood of the other team. In the meanwhile, agents in the same team cooperate with each other to establish communication protocol. Agents may take physical actions in the environment and do explicit communication actions that get broadcasted to other teammates. Each communication task consists of $M$ cooperative agents and each agent is capable of giving limited demonstrations to express its knowledge. 

In our setting, we do not assume that all agents have identical action and state space. Each agent observes the positions and the relative theta of other agents, and each agent's state space is represented by a 6-tuple vector, indicating the positions, theta, missiles, fuel and health. This vector space allows for $2.19 \times 10^{13}$ different states, and the corresponding action space consists of turning counterclockwise, turning clockwise, moving and firing.

\begin{figure}[ht]
\centering
\subfigure[Predictions of Categorical DQN at 20,000, 21,000, 40,000, 41,000 frames, respectively.]{
    \includegraphics[width=8cm]{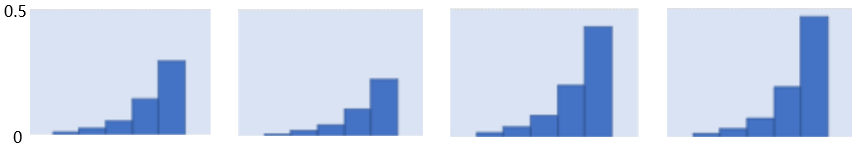}
}
\subfigure[Predictions of 9-1-LTCR at 20,000, 21,000, 40,000, 41,000 frames, respectively.]{
    \includegraphics[width=8cm]{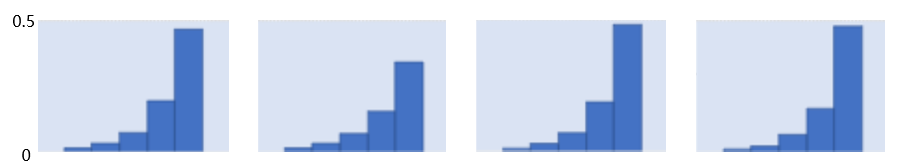}
}
\caption{In our analysis, we set the atoms $K$ to be 51, and we show the distributions at atoms 46-50 of independently learning (the first row) and 9-1-LTCR (the second row). We show the result at 20,000 and 40,000 frames, each with two sequential iterative distributions. It is noted that the probability distributions at 1-40 atoms are small enough to be ignore.}
\label{fig:distribute}
\end{figure}

\subsubsection{Parameter tuning}
As for our experiments, we set the learning rate to be $\alpha = 0.0001$, for the reason that the learning rate should be small enough to ensure convergence while applying model distillation. We use a simple $\epsilon$-greedy policy over the expected action-values. We set $\epsilon = 1$ at the start, and $\epsilon$ decreases to 0.01 gradually during the process of training, which indicates that agents are more inclined to choose actions output by the training networks rather than random actions, and we set the discount factor $\gamma = 0.99$. Notice that these parameters are consistent with previous research in these domains.

\subsection{Results and discussion}
In this section, we show our results of learning performance in single-player and multi-player mode. We also discuss why model distillation results in accelerating and stabilizing the  learning process with enhanced team-wide performance. 

\subsubsection{Single-player mode}
\textbf{Cartpole}
As for Cartpole, we set the total iteration rounds to be 100,000 frames. Furthermore, every 100 frames, we evaluate our agent's performance in Cartpole simulated environment with $\epsilon = 0.001$. To make comparisons convincing, we vary the ratios between the Revisit steps and Digest steps as 9:1, 7:3 and 5:5, namely 9-1-LTCR, 7-3-LTCR, 5-5-LTCR. The experimental results are moving averages over 500 frames.


\begin{figure}[ht]
    \centering
    \includegraphics[width=8cm]{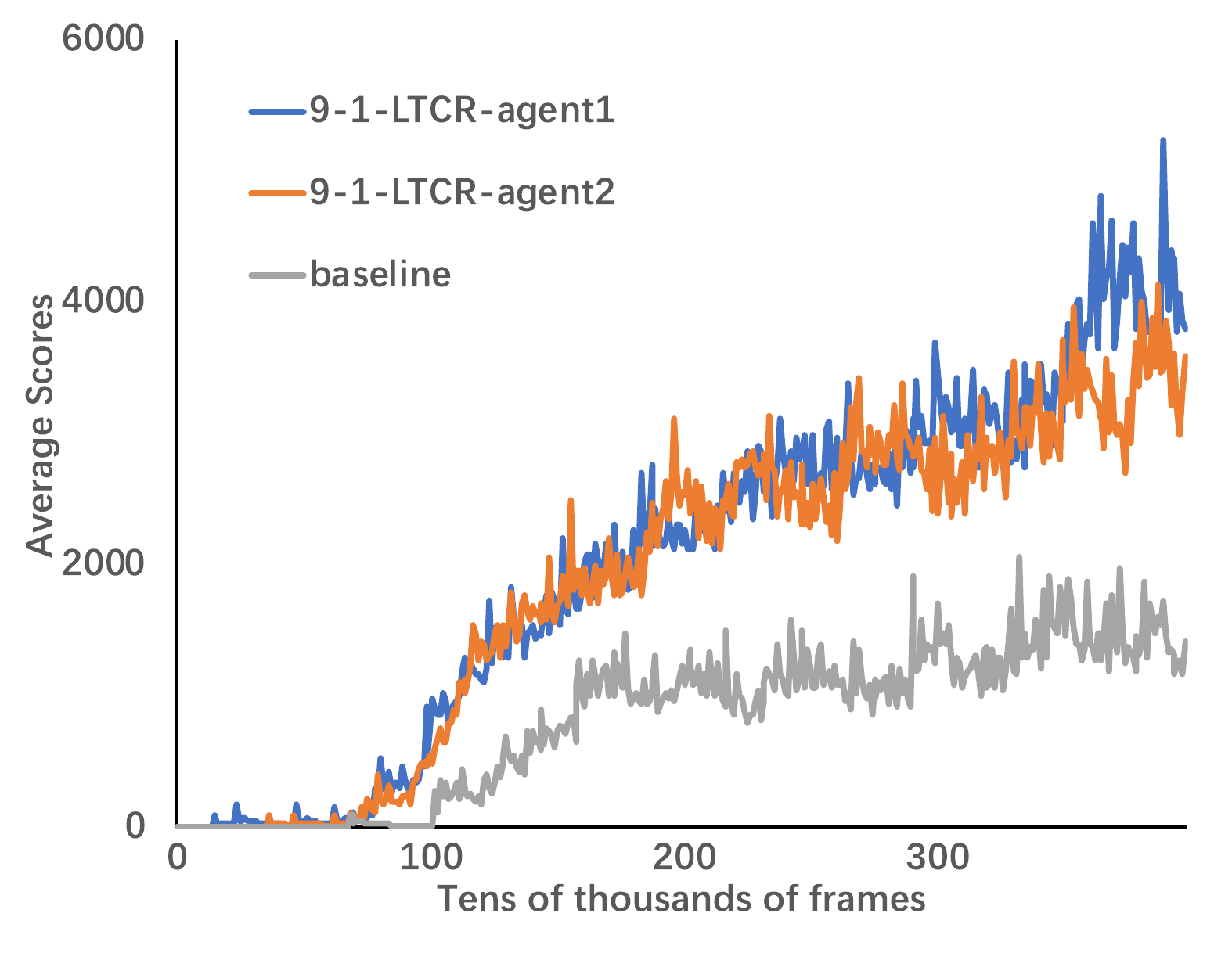}
    \caption{This figure indicates the learning curves in Zaxxon environment.}
    \label{fig:Zaxxon_reward}
\end{figure}

Figure \ref{fig:cartpoleReward} illustrates the tendency of agent's rewards. A first observation is that by the end of training, both agents are capable of optimal policy, whether with distillation or not. It is because Cartpole is a relatively simple game with a maximum score of 200, and all agents have learned a great strategy finally. We also note that by applying 9-1-LTCR among teammates, the rewards of two teammates increase much faster than that of those learning independently (the convergence speed is accelerated). Furthermore, the rewards received in the later stages of all LTCR training processes become more stable. It is because that target network of categorical DQN contains richer and more stable prediction of value functions than traditional DQN network. By distilling the output of categorical DQN, various predictions could be enhanced with heterogeneous knowledge, consequently with faster convergence. For arguing our analysis, we compare the distribution on the same feature vector at different training frames. As shown in Figure \ref{fig:distribute}, 
predictions trained by 9-1-LTCR are relatively stable and with smaller changes than baseline after 20,000 frames.
Obviously, model distillation accelerates the convergence of various predictions provided by stable targets.















    



Figure \ref{fig:cartpoleReward} also demonstrates that the curves of agent1 and agent2 basically coincide, for the reason that agent1's and agent2's value function distributions gradually converge to the same during the learning process. Therefore, 9-1-LTCR makes that the agents could explore separately while learning from teammates at the same time, finally with better performance. \\\\

\begin{table}[h]
\caption{Performance summary of Phoenix (The total iteration frames is 2,000,000 and then test agents' performance in Phoenix simulated environment every 5,000 frames. The maximum and average values are calculated by all recorded rewards.)}
\begin{tabular}{|c|c|c|}
\hline 
Methods         & Average Scores & Highest Scores \\ \hline \hline
beseline        & 156            & 1570           \\ \hline
9-1-LTCR-Agent1 & 184            & 2180           \\ \hline
9-1-LTCR-Agent2 & 167            & 2740           \\ \hline
\end{tabular}
\label{table:Phoenix}
\end{table}

\begin{table*}[ht]
\centering
\caption{Performance summary of space battle (the highest scores are also moving averages over 1000 frames).}
\begin{tabular}{|c|c|c|c|c|c|c|}
\hline
\multirow{2}{*}{Modes} & \multicolumn{3}{c|}{Average Scores}       & \multicolumn{3}{c|}{Highest Scores}       \\ \cline{2-7} 
                         & Learning independently & 9-1-LTCR & Gain  & Learning independently & 9-1-LTCR & Gain  \\ \hline \hline
2 vs 2                   & 6.75                   & 40.3     & \textbf{393\%} & 13.51                  & 55.6     & \textbf{312\%} \\ \hline
3 vs 3                   & 10.4                   & 35.3     & \textbf{239\%} & 17.9                   & 64.2     & \textbf{259\%} \\ \hline
4 vs 4                   & 8.75                   & 27.3     & \textbf{212\%} & 10.9                   & 36.2     & \textbf{232\%} \\ \hline
\end{tabular}
\label{table:spaceBattle}
\end{table*} 

\textbf{Arcade Learning Environment}
Such excellent performance of Cartpole could not be explained simply by the enhanced convergence speed of various predictions. As a result, we perform another two complex experiments.
As for Zaxxon, we set the total iteration rounds to be 4,000,000 frames and then test agents' performance in Zaxxon simulated environment with $\epsilon = 0.001$ every 10,000 frames.

Figure \ref{fig:Zaxxon_reward} and Table \ref{table:Phoenix} show the record of agent's reward of Zaxxon and Phoenix. We find that the rewards of the agents with model distillation are generally higher than the baseline with faster training speed, which is similar to Cartpole. So why we could obtain this performance? Does our method work well in multi-player mode? We will perform another experiment for further analysis.

\subsubsection{Multi-player mode}
\begin{figure}[ht]
\centering
\includegraphics[width=8cm]{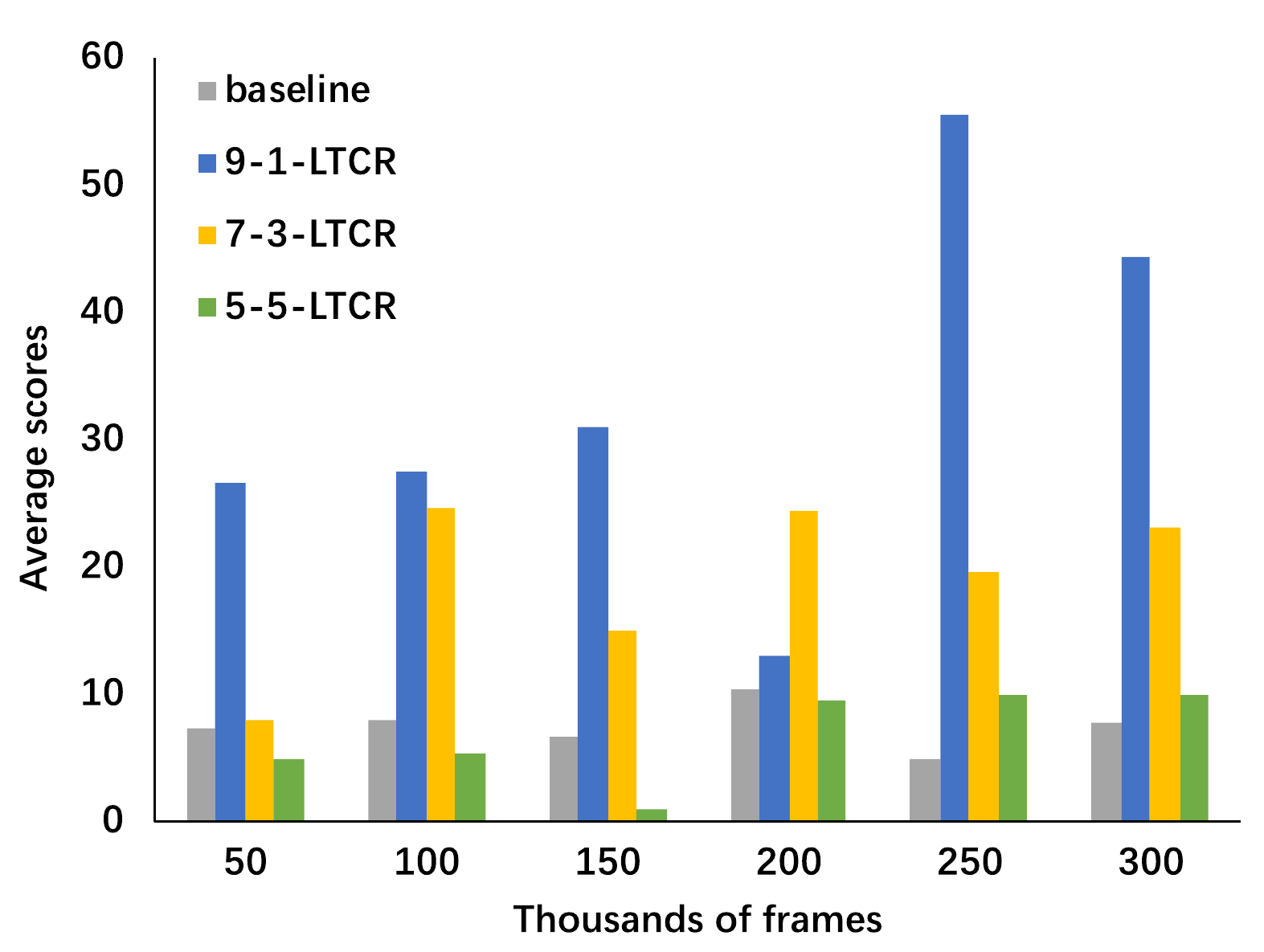}
\caption{This figure represents the learning process of Space Battle. And the experimental results are moving averages over 1000 frames.}
\label{fig:spaceBattleReward}
\end{figure}

\textbf{Space battle}
This section evaluates our methods in the Multi-player mode, showing the benefit of our methods in 1,000,000 frames. We also vary the ratios between Revisit steps and Digest steps as 9-1-LTCR, 7-3-LTCR, and 5-5-LTCR. We set the number of agents in one team to be 2. Furthermore, we show the results in 
Figure \ref{fig:spaceBattleReward}. We first examine the performance of traditional Categorical DQN. Despite the simple task, 
in practice we observe that the agent trained with traditional methods rarely shoot on the enemy, and we plot the learning progress over 300,000 frames. We must hypothesize that a primary reason for the failure in this setting is that the agents are with large state space but little action space. This problem is exacerbated as the number of timestamps grows: we observe traditional method are always results in worse rewards. This indicates that it is reasonable to adopt giving demonstrations to handle this environment.

Conversely, agents with LTCR could handle this environment more easily with model distillation. Figure \ref{fig:spaceBattleReward} illustrates the tendency of reward in 300,000 frames in the cooperative communication environment. As expected, our methods allow 9-1-LTCR outperforms any other approaches, and it turns out that all LTCR methods enhance performance compared to independent learning. We further set the agents in one team to be 3 and 4 shown in Table \ref{table:spaceBattle}. It is noted that 9-1-LTCR outperforms baseline more than 200\% in all modes. So where does the additional knowledge come from? Some intuitions about these questions are shown as follows: all students are primarily directed by the traditional loss function of Categorical DQN, which means that their performance generally increases. However, each network starts from different initial conditions. As a result, their probabilities estimates of next mostly likely vary, which provides extra information for better team-wide performance. Furthermore, another reliable interpretation of performance could also be made by visualizing \emph{dark knowledge} \cite{liu2019search}, which has already been investigated by some previous works about model distillation. It is commonly believed that the targets of a teacher can transfer \emph{dark knowledge} to enhance the student model. These results allow us to conclude the following truths:
\begin{enumerate}
    \item Our efficient demonstrations makes 9-1-LTCR outperforms in learning speed and time-wide performance because of \emph{dark knowledge}.
    \item LTCR is also scalable in multi-player mode since each agent is with limited action and state space.
\end{enumerate}

\section{Conclusion and Future Work}
This paper has introduced and evaluated LTCR, showing promising transfer between teammates in single-player and multi-player environments. Unlike prior works, our work does not assume expert teachers and does not need to follow or predict teammates' trajectories. Furthermore, we could explicitly represent the value functions of each agent that most methods learn. Results have shown that by applying different ratios between Revisit steps and Digest steps, we will obtain different performances. In our domain, 9-1-LTCR outperforms both the baseline and other LTCR methods. Such improvements result from three possible reasons:
\begin{enumerate}
    \item The stable target provided by Categorical DQN lays a foundation for accelerating learning.
    \item The rich predictions provided by Categorical DQN have been exploited by model distillation, which results in faster convergence.
    \item \emph{Dark knowledge} existing in model distillation further improves the team-wide performance than independent learning.
\end{enumerate}

Having shown the potential of LTCR, future works will consider a number of extensions. First, we will investigate the further performance of LTCR on more Atari 2600 games. Then, we will explore where does the \emph{dark knowledge} come from in deep reinforcement learning. 
\bibliographystyle{unsrt}
\bibliography{reff}
\end{document}